%% file: main.tex
\documentclass[a4paper,twoside]{article}
\usepackage{epsfig}
\usepackage{subcaption}
\usepackage{calc}
\usepackage{amssymb}
\usepackage{amstext}
\usepackage{amsmath}
\usepackage{amsthm}
\usepackage{multicol}
\usepackage{pslatex}
\usepackage{apalike}
\usepackage{algorithm2e}
\usepackage[bottom]{footmisc}
\usepackage{SCITEPRESS}    
\usepackage{hyperref}
\usepackage{booktabs}
\usepackage{graphicx}
\usepackage{mathtools}
\usepackage{subcaption}
\usepackage{todonotes}
\usepackage{multirow}
\usepackage{colortbl}
\usepackage{xspace}

\renewcommand{\thefootnote}{}

\usepackage{xcolor}
\usepackage{bm}
\usepackage{ulem}
\DeclarePairedDelimiterX{\infdivx}[2]{[}{]}{%
  #1\;\delimsize\|\;#2%
}
\newcommand{\infdiv}{\infdivx}

\newcommand{\approach}[0]{MPERL\xspace}

\newtheorem{definition}{Definition}
\begin{document}

\title{Markov Process-Based Graph Convolutional Networks for Entity Classification in Knowledge Graphs} 
\author{\authorname{Johannes M\"akelburg\sup{1*}\orcidAuthor{0009-0001-3821-7817},Yiwen Peng\sup{2*}\orcidAuthor{0009-0007-7902-4097},  Mehwish Alam\sup{2}\orcidAuthor{0000-0002-7867-6612}, Tobias Weller\sup{3} and Maribel Acosta\sup{1}\orcidAuthor{0000-0002-1209-2868}}
\affiliation{\sup{1}School of CIT, Technical University of Munich, Germany}
\affiliation{\sup{2}Télécom Paris, Institut Polytechnique de Paris, France}
\affiliation{\sup{3} Independent Researcher, Germany}
\email{\{johannes.maekelburg, maribel.acosta\}@tum.de, \{yiwen.peng, mehwish.alam\}@telecom-paris.fr}}
% Abstract
\input{tex/abstract}


\onecolumn \maketitle \normalsize \setcounter{footnote}{0} \vfill
\def\thefootnote{*}\footnotetext{These authors contributed equally to this work.}\def\thefootnote{\arabic{footnote}}

\input{tex/intro}

\input{tex/related_work}

\input{tex/rgcp}

\input{tex/experiments}

\input{tex/conclusion}

\vspace{-4mm}
\paragraph*{Acknowledgements:}  
This work has been funded by the Deutsche Forschungsgemeinschaft (DFG, German Research Foundation) - SFB 1608 - 501798263.

% ---- Bibliography ----
\bibliographystyle{apalike}
\bibliography{bibliography}

% ---- Appendix ----
%\input{appendix}

\end{document}

%% file: tex/abstract.tex
\abstract{Despite the vast amount of information encoded in Knowledge Graphs (KGs), information about the class affiliation of entities remains often incomplete. Graph Convolutional Networks (GCNs) have been shown to be effective predictors of complete information about the class affiliation of entities in KGs. However, these models do not learn the class affiliation of entities in KGs incorporating the complexity of the task, which negatively affects the models' prediction capabilities. To address this problem, we introduce a Markov process-based architecture into well-known GCN architectures. This end-to-end network learns the prediction of class affiliation of entities in KGs within a Markov process. The number of computational steps is learned during training using a geometric distribution. At the same time, the loss function combines insights from the field of evidential learning. 
The experiments show a performance improvement over existing models in several studied architectures and datasets. Based on the chosen hyperparameters for the geometric distribution, the expected number of computation steps can be adjusted to improve efficiency and accuracy during training.}

%% file: tex/intro.tex
\section{Introduction}

Knowledge Graphs (KGs) encode factual knowledge in the form of triples (subject-relation-object) and have emerged as a compelling abstraction for organizing semi-structured data, capturing relationships among entities. 
The facts available in KGs are used in many application areas such as recommendation~\cite{rec1}, information retrieval~\cite{ir1}, and question answering~\cite{qa} for improving the performance of these systems by providing background or auxiliary information. 
This relevance imposes a significant importance on a KG providing comprehensive information about the encoded entities.
In particular, encoding knowledge about the class affiliation of entities is of great importance for automatic reasoning and inferencing of information contained in a KG. 
Despite the enormous effort made to keep the knowledge encoded in the KGs up-to-date and consistent~\cite{heist2020knowledge}, KGs are often incomplete majorly due to automated constructions of KGs. 
To complete missing knowledge, in particular missing class affiliation of entities in KGs, various methods based on machine learning have been introduced~\cite{transe,gcn1}. 
Neural networks, especially Graph Convolutional Neural Networks (GCNs), have proven to be very effective in completing class affiliation of entities in KGs~\cite{rgcn}.
However, in neural network based methods, the computation cost grows with the size of the input data, but not with the complexity of the problem being learned. 
In recent developments in automated machine learning, models perform conditional computation based on probabilistic variables that are used to dynamically adjust the number of computation steps~\cite{ponder}. 
The adjustment of the computational budget, in particular the computation steps, is known as pondering.
Yet, existing graph-based machine learning algorithms do not consider the complexity of the task to adjust the number of computation steps to learn the parameters. 
%However, none of the existing graph-based machine learning algorithms take into account the complexity of the task and adjust the number of computation steps to learn the parameters. 
We address this issue and introduce a GCN-based model that learns to adapt the amount of computational steps based on the task at hand.

In this work, we propose Markov Process and Evidential with Regularization Loss (MPERL) that builds upon the previous idea of dynamically adjusting the number of computational steps based on the input of the model. We introduce an end-to-end Graph Convolutional Network (GCN) model that is learned within a Markov process and use recent developments in the field of evidential learning~\cite{evidential_learning1,el2}.
Previous work has demonstrated the high performance of evidence-based models. 
Unlike models using a softmax function, evidence-based models are effective predictors that do not make overconfident predictions. 
We therefore follow an evidence-based approach as well. 
The Markov process in which the model is learned consists of two states: (i) the \textit{continue} state, indicating further computational steps, and (ii) the \textit{halt} state, indicating the end of the computational steps.  
The overall probability of halting at each step is modeled as a geometric distribution. Unlike previous work for entity classification in KGs, MPERL is a graph-based model that dynamically adjusts the number of computational steps according to complexity. 
MPERL further shows its feasibility on the task of entity type prediction, i.e., inferring the knowledge about the class affiliation of an entity. 
In the rest of this paper, we are treating entity type prediction as a classification task where we perform single-label classification for smaller datasets (specifically designed for this purpose) and multi-label classification on larger datasets. 
The experimental results show that MPERL (GCN with markov process and evidential loss) outperforms vanilla GCN as well as various other baselines.
The ablation studies show that the use of both markov process and the evidential learning loss provide significant increase in the performance of the MPERL. Overall our paper makes the following contributions:

\begin{itemize}
    \item[$\bullet$] We introduce a Graph Convolutional Network based model trained within a Markov process, using an evidential loss function.
    \item[$\bullet$] We demonstrate the performance of the model in predicting missing class affiliations of entities using single- and multi-label classification.
    \item[$\bullet$] We show the effect on the model when adapting the number of computational or Markov steps.
    \item[$\bullet$] We show the effectiveness of each of the components of the model on the overall results with the help of an ablation study. 
\end{itemize}

The paper is structured as follows: Section~\ref{sec:related_work} discusses the recent works related to representation learning over KGs for entity classification. Section~\ref{sec:approach} details the proposed approach while Section~\ref{sec:experiments} shows the effectiveness of MPERL with the help of thorough experimentation over various sizes of the datasets as well as the ablation study. Finally, Section~\ref{sec:conclusions} concludes the study and discusses future directions.

%% file: tex/related_work.tex
\section{Related Work}
\label{sec:related_work}

Different learning approaches have been applied to the problem of entity classification in KGs~\cite{alamtowards}. 
Relational Graph Convolutional Networks (R-GCN)~\cite{rgcn} uses the structure of KGs to generate embeddings based on local neighbors in order to predict classes~\cite{gcn1,graphsage}. Due to their strong performance on graph-structured data, GCN models have been particularly used and extended in recent years to tackle entity classification~\cite{rgcn,GatedRGCN,attentionentityclassification}, relation classification~\cite{relationgcn}, and KG alignment~\cite{kgAlignment1,kgAlignment2}. Gated Relational Graph Neural Network (GRGNN)~\cite{GatedRGCN} introduced a gate mechanism to leverage hidden states of current node and its neighbors to target entity classification problem in KGs. 
Whereas Relational Graph Attention Networks~\cite{rgat} and Multilayer Graph Attention Network~\cite{attentionentityclassification} use masked self-attentional layers to learn the weighting factor of neighboring node's features and were extended with intra- and inter-layer connections between nodes.
Evidential Relational-Graph Convolutional Networks (\textit{E}-R-GCN)~\cite{ergcn} extend R-GCN~\cite{rgcn} with an evidential loss to represent the predictions of the model as a distribution over possible softmax outputs and estimate the associated evidence to learn both aleatory and epistemic uncertainty in entity classification.
In contrast to these approaches, our work also implements a Markov process to learn the model. 
Moreover, translational KG embeddings (e.g, TransE~\cite{transe} and extensions) and factorization-based KG embeddings (e.g., DistMult~\cite{distmult} and RESCAL~\cite{rescal}) have been proposed.
In general, these embeddings are particularly effective for link prediction, but less for entity classification~\cite{class}. TransET~\cite{TransET} is an extension of TransE~\cite{transe} that implements a convolution-based projection of entities into a type-specific representation to address entity classification. 
ConnectE~\cite{connectE} is also a translational-based approach that learns two distinct embedding models of the entities and connects them via a joint model to predict entity types. Ridle~\cite{ridle} computes a distribution over the use of relations of entities using a stochastic factorization model.
Besides translational and factorization-based embeddings, RDF2Vec~\cite{rdf2vec} generates a sequence of nodes using random walks and Weisfeiler-Lehman subtree RDF graph kernels that are passed to Word2Vec language model for learning low-dimensional numerical representations of entities. The learned embeddings preserve similar entities closer in the vector space, which makes RDF2Vec suitable for entity classification~\cite{rdf2vec_type1,rdf2vec_type2,rdf2vec_type3}. 
Our solution differs from these approaches in the combination of evidential learning with a Markov process. This allows our approach to learn  embeddings tailored to entity classification while adjusting the number of computational steps according to the complexity of the task at hand.

Other KG representations for entity classification have also been proposed which utilize semantic information related to an entity.  Cat2Type~\cite{cat2type} creates representations for entities based on the textual information available in the Wikipedia category names using language models and the category network information.
In addition to textual information related to entities, GRAND~\cite{BiswasPPSA22} uses several kinds of graphs such as entity based, relation based, and random walks for considering the strcutured contextual information of an entity.  
In~\cite{esws/RiazAG23}, the authors perform entity typing based only on the labels as well as descriptions of the entities using BERT-based models.
These approaches can only be applied to KGs where class affiliation can be predicted by the relation distribution~\cite{ridle} or where additional semantic information is available. 
Relational aggregation graph attention network (RACE2T)~\cite{zou2022knowledge} proposes a method consisting of an encoder which consists of the attention coefficient between entities further used to aggregate the information of relations and entities in the neighborhood of the entity. The decoder is based on a convolutional neural network. 
Lastly, ASSET~\cite{asset} is a semi-supervised approach that learns from massive unlabeled data for entity classification. Compared to our work and the related work above, ASSET does not learn embeddings itself, but uses existing ConnectE~\cite{connectE} embeddings learned beforehand on the KG.

%% file: tex/rgcp.tex
\section{Our Approach: MPERL}
\label{sec:approach}

In this section, we introduce our method Markov Process and Evidential with Regularization Loss (MPERL), that extends current Graph Convolutional Networks (GCN) to perform entity classification in KGs. 
For this purpose, we first introduce the definition of a KG and the associated research problem.
\begin{definition}
A Knowledge Graph $\mathcal{KG}$ is a tuple $(\mathcal{E}, \mathcal{R}, \mathcal{L}, \mathcal{C})$, where the pair-wise disjoint sets $\mathcal{E}$, $\mathcal{R}$, $\mathcal{L}$, and $\mathcal{C}$ correspond to the set of entities, relations, literals, and types or classes, respectively. A statement in $\mathcal{KG}$ is modelled as a triple $(s,r,o)$, with $s \in \mathcal{E} \cup \mathcal{R} \cup \mathcal{C}$, $r \in \mathcal{R}$, and $o \in \mathcal{E} \cup \mathcal{R} \cup \mathcal{L} \cup \mathcal{C}$. 
\end{definition}   

The problem of entity classification in a $\mathcal{KG}$ is to predict statements $(e,r,C)$ that should be in $\mathcal{KG}$, where $e \in \mathcal{E}$, $r \in \mathcal{R}$ denotes the class affiliation relationship, and $C \in \mathcal{C}$ is a class. To address this problem, we present both the architecture and the learning process of \approach in Section~\ref{sec:ponder}. In Section~\ref{sec:loss}, the loss function for learning the parameters of the model is presented.

\subsection{Markov Process Extensions for Entity Classification}
\label{sec:ponder}

\begin{figure*}[t!]
\includegraphics[width=0.98\textwidth]{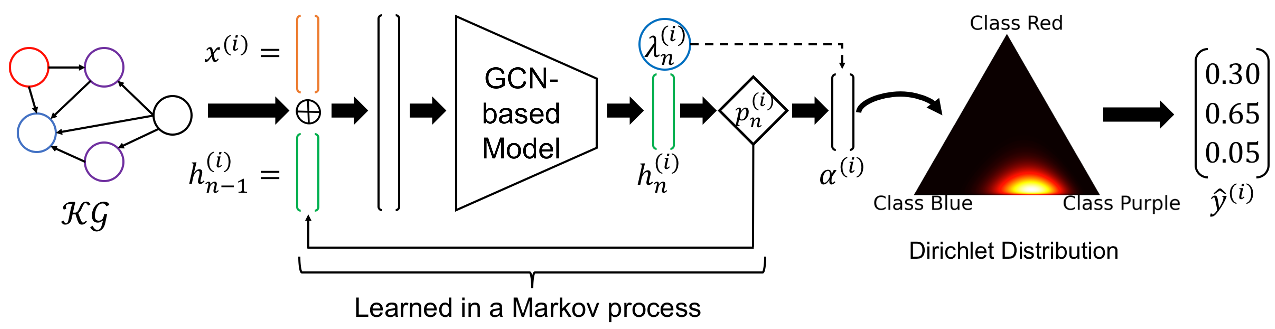}
\caption{Approach for entity classification using \approach. \approach gets as input a one-hot encoding of the entity ID (denoted $x^{(i)}$) and the learned hidden features from the previous Markov step.  
The GCN-based model uses the structure of the KG to compute the hidden features $h^{(i)}_{n}$ and the halting probability $\lambda_{n}^{(i)}$. The prediction $\hat{y}^{(i)}$ is based on the Dirichlet parameters $\alpha^{(i)}$.}
\label{fig:architecture}
\end{figure*}

\textbf{Overview.} 
We model the entity classification problem as a supervised learning problem. 
Figure~\ref{fig:architecture} shows the overall architecture of our proposed solution \approach, which integrates a Markov process into a GCN-based model, e.g., R-GCN~\cite{rgcn}. 
First, a representation of the entities in the KG is learned (cf. Eqs.~1-3), which relies on GCNs to represent entities from the KG and are used in each step of the Markov process to calculate the hidden layers. The Markov process in which the model is learned is defined in Eqs.~4-6.  
We use a generalized geometric distribution to model the transition probabilities of the two states (\textit{halt} and \textit{continue}) of the Markov process. We learn with parameter $\lambda_n^{(i)}$ (cf. Eq.~5) a parameter from which we can derive the probability in which step of the Markov process the halt state is reached (cf. Eq.~6). 
By learning this parameter, the number of Markov steps and, thus, the number of epochs is adapted based on the input of the model. The final output of \approach is given in Eqs.~7-9. In Eq.~7, the features of the individual steps of the Markov process are aggregated by weighted means and used to parameterize a Dirichlet distribution (Eq.~8). 
We use a Dirichlet distribution since this is the only conjugate prior for a categorical distribution used to indicate the probability of class affiliation of an entity in a KG. The prediction of a sample $i$ is described by the expected probability of the Dirichlet distribution in Eq.~9. We use the expected probability as prediction for entity types due to its property of unbiased manners.

\smallbreak
\noindent
\textbf{Learning Process.} 
For each entity $e^{(i)} \in \mathcal{E}$  of the KG $\mathcal{KG}$ we initialize each entity representation by concatenating one hot encoding with hidden state of previous markov step. 
We denote this vector as $x^{(i)}$. This vector is concatenated with the hidden feature representation of the entity $e^{(i)}$ of the previous Markov step denoted as $h^{(i)}_{n-1}$. 
Initially, in step $n=1$, $h^{(i)}_{0}$ is a null vector, so the feature representation is $h^{(i)}_{0} = \overrightarrow{0}$. $n \in [1,N]$ denotes the current step of the Markov process where $N$ is the maximum number of Markov steps. The concatenation of the two vectors $x^{(i)}$ and $h^{(i)}_{n-1}$ is used as input to the neural network in step $n$ in the Markov process. We denote the neural network input, i.e., in layer $l=0$, of a sample $i$ in Markov step $n$ as follows:

\begin{equation}
h^{(i)[0]}_{n} = \infdiv{x^{(i)}}{h^{(i)}_{n-1}}
\label{eq:first_layer}
\end{equation}
where $\infdiv{}{}$ is the concatenation operation and the number in square brackets in superscript denotes the considered layer.

By incorporating the hidden feature representation $h^{(i)}_{n-1}$ from the previous step, the learned features are reused to enable faster convergence. 
The fundamental concept is similar to GCRNN, although rather than using $h^{(i)}_{n-1}$ as the only input to \approach, the concatenation of $x^{(i)}$ and $h^{(i)}_{n-1}$ is used to avoid overfitting. 

The hidden representation of an entity ${e^{(i)} \in \mathcal{E}}$ in layer $l+1$ is then computed using a simple propagation model to calculate the forward pass update. 
For updating the entity representation, we apply full sampling or partial neighbourhood sampling (for larger datasets) during message passing phase in graph neural networks.
\begin{equation}
h_n^{(i)[l+1]} = \phi\left(\sum_{r \in \mathcal{R}}\sum_{j \in \mathcal{N}_i^{r}} \dfrac{1}{|\mathcal{N}_i^r|}W_{r}^{[l]}h_n^{(j)[l]} + W_0^{[l]}h_n^{(i)[l]} \right)
\label{eq:hidden_layer}
\end{equation}

$\phi$ denotes the ReLU function and $\mathcal{N}_i^{r}$ denotes the indices of neighboring nodes with relation $r \in \mathcal{R}$ to the node with index $i$. $l \in [1,L]$ denotes the layer with $L$ as the number of layers in the neural network. In Eq.~\ref{eq:hidden_layer}, the feature representations of the neighboring nodes of the node with index $i$ are relation-specifically aggregated with the weight matrix $W_{r}^{[l]}$ and normalized by the number of neighboring nodes ($\frac{1}{|\mathcal{N}_i^r|}$). 
This relation-specific transformation is summed up and extended by a self-loop to include the current representation of the node itself. The ReLU function $\phi$ is applied as a non-linear activation function.
For regularizing the network layers' weights, basis decomposition is used to avoid a rapid growth in the number of parameters with the number of relations in the graph.
Basis decomposition uses a linear combination of basis transformations $V_b^{[l]} \in \mathbb{R}^{d^{[l+1]} \times d^{[l]}}$ with coefficients $a_{rb}^{[l]}$ such that only the coefficients depend on $r$.
\begin{equation}
W_r^{[l]} = \sum_{b=1}^{B}a_{rb}^{[l]}V_b^{[l]}
\label{eq:basis_decomposition}
\end{equation}

For ease of reading, we denote the hidden representation of the last layer of sample $i$ in step $n$ as $h_n^{(i)}$.
\approach learns a function $\mathcal{S}(x^{(i)},h^{(i)}_{n-1})$ that outputs the parameters of a Dirichlet distribution $\alpha^{(i)}$, used as conjugate prior of a categorical distribution from which the predictions $\hat{y}^{(i)}$ are drawn, the hidden features $h^{(i)}_{n}$ and the probability of halting $\lambda^{(i)}_{n}$ at current step.
The function $\mathcal{S}(x^{(i)},h^{(i)}_{n-1})$ is learned within a Markov process. We use $\lambda_{n}^{(i)}$ to learn the optimal value $n$. The Markov process uses a Bernoulli random variable, which we denote as $\Lambda_{n}$, to represent the two states \textit{continue} ($\Lambda_{n} = 0$) and \textit{halt} ($\Lambda_{n}=1$). \textit{halt} is an absorbing state, meaning that, once entered, cannot be left. This defines the end of learning within the Markov process. The Markov process initially starts in the \textit{continue} state, therefore $\Lambda_{0}=0$ holds. The transition probability of a sample $i$ that the state \textit{halt} is assumed in step $n$, given that the previous step was \textit{continue} is expressed by the following conditional probability:
\begin{equation}
P(\Lambda_{n}=1 | \Lambda_{n-1}=0) = \lambda_{n}^{(i)} \quad \forall \: 1 \leq n \leq N
\end{equation}

The conditional probability $\lambda_{n}^{(i)}$ is computed using a sigmoid function $\sigma$ with parameters $U \in \mathbb{R}^{d^{[L]} \times 1}$ and $h_{n}^{(i)} \in \mathbb{R}^{d^{[L]} \times 1}$, where $d^{[L]}$ denotes the number of dimensions of the hidden features of $h^{(i)}_{n}$ in the last layer.
%
% Predictions made
\begin{equation}
\lambda_{n}^{(i)} = \sigma(h^{(i)}_{n}) = \frac{\mathrm{1} }{\mathrm{1} + e^{- U^{T}h^{(i)}_{n} }}
\end{equation}

The probability of entering the state \textit{halt} in step $n \in [1,N]$ can be derived by means of the following generalized geometric distribution $p_{n}$.
\begin{equation}
p_{n}^{(i)} = \lambda_{n}^{(i)}\prod^{n-1}_{s=1}(1-\lambda_{s}^{(i)})
\end{equation}

$p_{n}^{(i)}$ defines for sample (i.e., entity) $i$ the probability of entering the absorbing state $\Lambda_{n}=1$ for the first time in step $n$, based on $\lambda_{n}^{(i)}$.

Once the absorbing state $\Lambda_{n}=1$ has been entered, the Markov process terminates and the learned hidden features of each step, $h^{(i)}_{s}$ with $1 \leq s \leq n$, are aggregated. 
In contrast to existing work, which uses the final output $h^{(i)}_n$~\cite{ponder} or the weighted average across all steps for prediction ($\sum^{n}_{s=1}\hat{y}_{s} \lambda_{s}$), we follow a different approach and use a weighted average of the hidden features across all steps as final hidden feature. 
\begin{equation}
h^{(i)} = \sum^{n}_{s=1}h_{s}^{(i)} \lambda_{s}^{(i)}
\label{eq:hidden_feature}
\end{equation}

In Eq.~\ref{eq:hidden_feature}, $\lambda_{s}^{(i)}$ denotes for sample $i$ the probability to enter the absorbing state in step $s$. A high $\lambda_{s}^{(i)}$ value, with $1 \leq s \leq n$, is due to a fitting feature representation $h_s^{(i)}$ to predict $\hat{y}^{(i)}$, thus, a high significance is assigned to this feature representation when aggregating $h^{(i)}$.\looseness=-1

$h^{(i)} \in \mathbb{R}^{d^{[L]}}$ is thus an aggregation of the features of the individual steps in the Markov process and is used as parameter of the conjugate prior. 
The general idea is that no softmax function is used to predict the categorical values, but a conjugate prior categorical distribution from which the predictions are drawn. 
The advantage over a softmax function is that not just one point estimator is available for prediction, but a large number of categorical distributions that can be drawn from the conjugate prior. 
At the same time, the uncertainty can be quantified by the conjugate prior. Given the supervised learning problem for predicting categorical values $y^{(i)} \in \mathcal{C}$, where $\mathcal{C}$ is the set of classes in $\mathcal{KG}$ and $K$ denotes the number of classes (i.e. $\mathcal{C} = \{C_{1}, \dots , C_{K}\}$), we use a Dirichlet distribution as conjugate prior of a categorical distribution.
Depending on the predictions of the Dirichlet distribution's parameters, the concentration of the drawn distributions can be on one class or, if  uncertainty is large, it can be spread over several classes.

In order to determine the Dirichlet parameters $\alpha^{(i)} \in \mathbb{R}^{K}_{+}$ for sample $i$, where $K\geq2$ always holds, we use the aggregated hidden feature representation $h^{(i)}$ (see Eq.~\ref{eq:hidden_feature}) and a ReLU function $\phi$ to determine the parameters of the Dirichlet distribution as follows:

\begin{equation}
  \alpha^{(i)} = \phi(h^{(i)}) + 1
  \label{eq:alpha}
\end{equation}

The ReLU function $\phi$ outputs values in the range $[0,\infty)$. Since we add 1 in the Eq.~\ref{eq:alpha}, the constraint of the Dirichlet parameter $\alpha^{(i)} \in \mathbb{R}^{K}_{>1}$ holds.
The prediction $\hat{y}^{(i)}$ of a sample is the expected probability of the Dirichlet distribution with parameter $\alpha^{(i)}$.
\begin{equation}
  \hat{y}^{(i)} = \frac{\alpha^{(i)}}{\sum^K_{k=1} \alpha_k^{(i)}}
  \label{eq:prediction}
\end{equation}

\subsection{Evidential with Regularization Loss}
\label{sec:loss}

Based on existing work~\cite{ergcn}, we have chosen to use an evidential loss function rather than a cross-entropy function~\cite{ponder,rgcn}. 
However, to simultaneously control the number of steps within the Markov process, our loss function $L$ consists of two terms $L_{ev}$ and $L_{reg}$. 
The evidential loss $L_{Ev}$ optimizes the parameter for fitting the predictions $\hat{y}$ to the target values $y$, and the regularization loss $L_{Reg}$ optimizes the parameter for the number of Markov steps. 
For the sake of readability, the following equation defines the loss function for one sample.% \\

The loss function $L$ combines fundamental concepts of \textit{E}-R-GCN~\cite{ergcn}, PonderNet~\cite{ponder}, and uncertainty quantification in neural networks~\cite{evidential_learning1}. The loss $L$ and the corresponding adjustment of the parameters of the model is computed and adjusted after each epoch and not after each step of the Markov process. The reason for this is that if the weights are adjusted after each Markov step, the rates of convergence are lower, because the network adjusts itself in each Markov step and, thus, produces volatile results. 
In contrast, computing the loss and adjusting the parameters of the model after each epoch is more natural and allows  smoother convergence of the parameters. 

\begin{equation}
\begin{aligned}
    L = & p_n \left\{ \sum_{k=1}^{K} \Bigg( 
        \underbrace{(y_{k} - \hat{y}_{k})^2}_{L_{Ev}^{err}} 
        + \underbrace{\frac{\hat{y}_{k}(1-\hat{y}_{k})}{\sum_{k=1}^K \alpha_k + 1}}_{L_{Ev}^{var}} 
        \Bigg) \right.  \\
    &\left. + \delta_t \underbrace{KL\left(D(\tilde{\mathbf{\alpha}})\,||\,D(\langle 1, \dots, 1 \rangle)\right)}_{L_{Ev}^{unc}} \right\} 
     {L_{Ev}} \\
    &+ \beta \underbrace{KL\left(p_{n}\,||\,p_{G}(\lambda_{p})\right)}_{\text{Regularization loss } L_{Reg}}
\label{eq:loss}
\end{aligned}
\end{equation}

In our loss function, $L_{Ev}$ is the evidential loss across halting steps. Consistent with previous work in evidential learning, the evidential loss $L_{Ev}$ consists of three components: minimizing the error of prediction $\hat{y}$ ($L_{Ev}^{err}$), minimizing the variance of the Dirichlet distribution to reduce uncertainty ($L_{Ev}^{var}$), and a regularization term which penalizes the predictive distribution, which does not contribute to data fit ($L_{Ev}^{unc}$). In order to ensure that the evidential loss is stable even for samples that do not follow the predicted distribution and, therefore, cannot be correctly classified but the loss still decreases towards zero, the Kullback-Leibler (KL) divergence is built into the evidential loss $L_{ev}$. 
In related work, it has been shown that using the KL divergence for out-of-distribution samples provides more stable performance in prediction~\cite{evidential_learning1,ergcn}, which is why we also use it in our loss $L_{Ev}$ and define it as $L_{Ev}^{unc}$ in Eq.~\ref{eq:loss}. 
$L_{Ev}^{unc}$ is multiplied by an annealing coefficient $\delta_{t} = min(1.0, t/10) \in [0, 1]$. We gradually increase this coefficient within the first 10 epochs and keep it fixed afterwards to ensure that the influence of the annealing coefficient increases over the epochs but does not exceed. In this way, we prevent an early convergence to a uniform distribution for the misclassified samples and allow the network to explore the parameter space at the beginning.

In $L_{Ev}^{unc}$, $D(\tilde{\mathbf{\alpha}})$ denotes the Dirichlet distribution with parameter $\tilde{\mathbf{\alpha}}$ and $D(\langle 1,\dots ,1 \rangle)$ denotes the uniform Dirichlet distribution. $\tilde{\mathbf{\alpha}}$ is the adjusted evidence of the previous parameter $\alpha$ and is defined as follows.

\begin{equation}
\tilde{\mathbf{\alpha}} = y + (1 - y) \alpha
\end{equation}

The regularization term $L_{Ev}^{unc}$ of the evidential loss with annealing coefficient $\delta_{t}$, epoch $t$, gamma function $\Gamma(\cdot)$ and digamma function $\psi(\cdot)$ is as follows.

The second term of the loss function $L$ (see Eq.~\ref{eq:loss}) is the regularization loss $L_{Reg}$. 
$L_{Reg}$ uses the Kullback-Leibler (KL) divergence to measure the difference between the distribution of halting probabilities $p_{n}$ at step $n$ and a prior geometric distribution denoted as $p_{G}(\lambda_{p})$.
The reason for using the regularization loss $L_{Reg}$ is that it may improve generalization. 
In addition, it provides an incentive to keep the number of Markov steps performed no longer than the given distribution $p_{G}(\lambda_{p})$. 

\begin{equation}
\begin{multlined}
KL\left(D(\tilde{\mathbf{\alpha}})\,||\,D(\langle 1, \dots, 1 \rangle)\right) 
    = \\ 
     \hspace{-6mm} \log\left( \frac{\Gamma \left(\sum_{k=1}^{K}\tilde{\mathbf{\alpha}}_{k}\right)}{\Gamma(K) \prod_{k=1}^{K} \Gamma(\tilde{\mathbf{\alpha}}_{k})} \right)   \\
    \ \ \ \quad + \sum_{k=1}^{K} (\tilde{\mathbf{\alpha}}_{k}-1) \left[ \psi(\tilde{\mathbf{\alpha}}_{k}) - \psi\left(\sum_{j=1}^{K}\tilde{\mathbf{\alpha}}_{j}\right) \right]  
\end{multlined}
\end{equation}

The goal of KL in the regularization term $L_{Reg}$ is to approximate the distribution of $p_{n}$ to the geometric prior probability distribution $p_{G}(\lambda_{p})$, which is defined by the hyperparameter $\lambda_{p}$.
This distribution describes the probability that the model enters the absorbing state ($\Lambda_n=1$) in step $n$  as follows.

\begin{equation}
p_{G}(\lambda_{p}) = (1 - \lambda_{p})^{n}\lambda_{p}
\label{eq:geometric}
\end{equation}

Using the geometric prior probability distribution $p_{G}(\lambda_{p})$, an incentive is given to the network to approximate the number of Markov steps to the expected value of the geometric prior probability distribution $\mathbb{E}(p_{G}(\lambda_{p})) = \frac{1}{\lambda_{p}}$, i.e. promotes exploration. 
This incentive can be controlled by the hyperparameter $\beta$ and is $0.01$ in our study. $L_{Reg}$ in Eq.~\ref{eq:loss} is defined as follows.

\begin{equation}
KL(p_{n}||p_{G}(\lambda_{p})) = \log\Bigg(\frac{\lambda_{n}}{\lambda_{p}}\Bigg) + \frac{1}{\lambda_{n}}\log\Bigg(\frac{1-\lambda_{n}}{1-\lambda_{p}}\Bigg)
\end{equation}

In summary, the loss function $L$ (see Eq.~\ref{eq:loss}) thus has two functions. On the one hand, the conjugate prior, which in our case is a Dirichlet distribution, is to be fitted in such a way that the deviations between the target values $y$ and the predictions $\hat{y}$ are minimized by $L_{Ev}$. And on the other hand, the number of Markov steps should be controlled by $L_{Reg}$.

%% file: tex/experiments.tex
\section{Experiments}
\label{sec:experiments}
First, we provide the experimental configuration (\S\ref{sec:experimental_setup}). 
In our experimental study, we investigate the following questions: 
\textbf{(Q1)} How effective is \approach on state-of-the-art benchmarks? (\S\ref{sec:accuracyresults}) 
\textbf{(Q2)} What are the effects of the hyperparameter $\lambda_p$ of \approach? (\S\ref{sec:iterations})
\textbf{(Q3)} What is the impact of the \approach components on the performance? (\S\ref{sec:ablation}) 
The source code and the datasets are available online\footnote{\url{https://github.com/DE-TUM/MPERL}}.

\begin{table}[t!]
\setlength{\tabcolsep}{2pt}
      \caption{Dataset statistics}
    \centering
      \label{tab:dataset}
      \scriptsize
      \begin{tabular}{lrrrrr}
    \toprule
    \textbf{Datasets} & Entities & Relations & Triples & Labelled & Classes \\
    \midrule
    \textbf{AIFB} & 8,285 & 45 & 29,043 & 176 & 4 \\
    \textbf{MUTAG} & 23,644 & 23 & 74,227 & 340 & 2 \\
    \textbf{BGS} & 333,845 & 103 & 916,199 & 146 & 2 \\
    \textbf{AM} & 1,666,764 & 133 & 5,988,321 & 1,000 & 11 \\
    \midrule
    \textbf{FB15kET} & 14,951 & 1,345 & 483,142  & 168,313 & 3,584 \\
    \textbf{YAGO43kET} & 42,335 & 37 & 331,686 & 462,083 & 45,182 \\
    \bottomrule
\end{tabular}
\end{table}

\subsection{Experimental Setup}
\label{sec:experimental_setup}
\begin{figure}[t!]
    \centering
    \begin{subfigure}[b]{0.15\textwidth}
        \includegraphics[width=\textwidth]{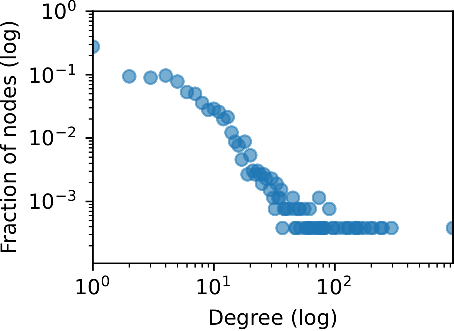}
        \caption{AIFB}
        \label{fig:aifb}
    \end{subfigure}%
    \begin{subfigure}[b]{0.15\textwidth}
        \includegraphics[width=\textwidth]{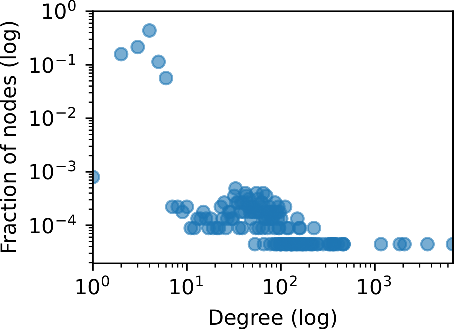}
        \caption{MUTAG}
        \label{fig:mutag}
    \end{subfigure}%
    \begin{subfigure}[b]{0.15\textwidth}
        \includegraphics[width=\textwidth]{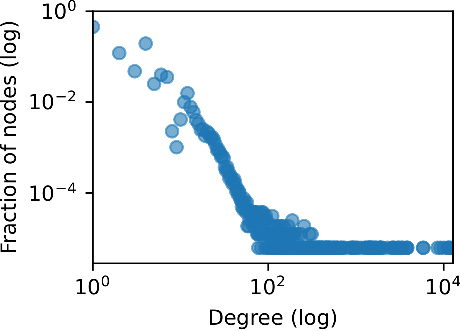}
        \caption{BGS}
        \label{fig:bgs}
    \end{subfigure}%
    \\[1ex] 
    \begin{subfigure}[b]{0.15\textwidth}
        \includegraphics[width=\textwidth]{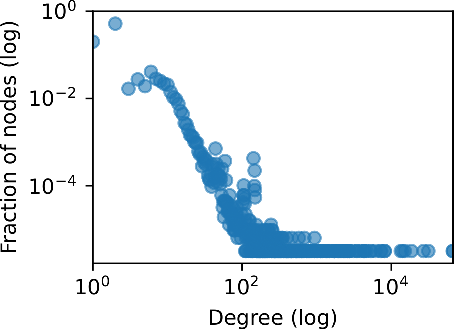}
        \caption{AM}
        \label{fig:am}
    \end{subfigure}%
    \begin{subfigure}[b]{0.15\textwidth}
        \includegraphics[width=\textwidth]{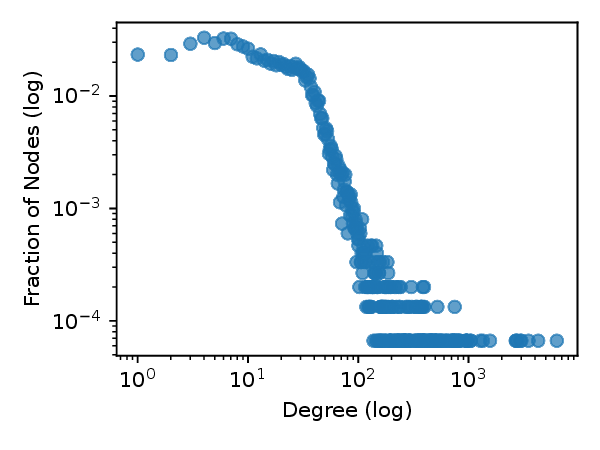}
        \caption{FB15kET}
        \label{fig:fb}
    \end{subfigure}%
    \begin{subfigure}[b]{0.15\textwidth}
        \includegraphics[width=\textwidth]{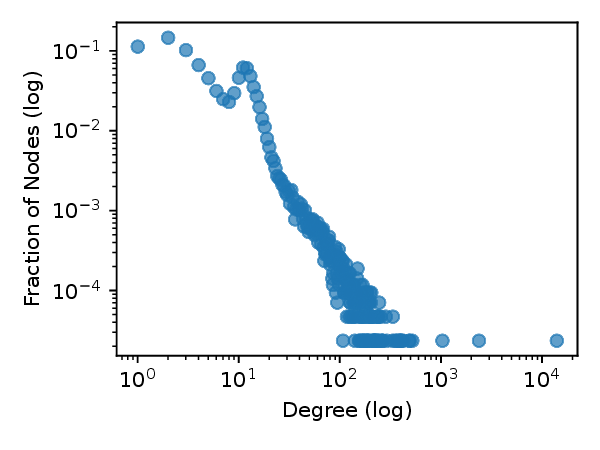}
        \caption{YAGO43kET}
        \label{fig:yago}
    \end{subfigure}
    \caption{Degree distribution of entities in datasets}
    \label{fig:degrees}
\end{figure}

\textbf{Datasets.}
The evaluation is performed using the standard SOTA datasets used for entity classification, i.e., AIFB, MUTAG, BGS, and AM for evaluation~\cite{ristoski2016collection}. 
In AIFB, the class affiliation is modeled by the relation \texttt{employs} and \texttt{affiliation}, MUTAG by \texttt{isMutagenic}, BGS by \texttt{hasLithogenesis}, and AM by \texttt{material}. 
The triples containing these relations have been removed from training.
We use predefined train/test splits, which are provided with the datasets.  
In addition to these benchmark datasets we consider two additional larger benchmarks derived from real-world knowledge graphs, i.e., FB15kET~\cite{bordes2013translating} and YAGO43kET~\cite{moon2017learning}. 
We follow the proposed train/valid/test split. 
The dataset statistics are summarized in Table~\ref{tab:dataset},  and the degree of distribution among the entities in the graphs is shown in Figure~\ref{fig:degrees}.
\smallbreak

\noindent
\textbf{Metrics.}
We report on the accuracy and F1-macro score for the four smaller datasets. For the larger datasets, we use two ranking-based metrics in a filtered setting\footnote{Following~\cite{bordes2013translating}, filtered setting means that all the known types of entity $e$ in the training, validation, and test sets are first removed from the ranking, allowing us to obtain the exact rank of the correct type $t_e$ among all types.}: 
Mean Reciprocal Rank (MRR) and proportion of correct entity types predicted in top k (HIT@$k$, $k=1, 3, 10$).
Each experiment is run ten times, and we present the average performance over the test splits. 

\smallbreak
\noindent
\textbf{Approaches.}
We implemented \approach using R-GCN as the GCN-based model in the architecture. Therefore, our experiments report on MPERL+R-GCN as our approach.  
The baselines used in the experiments include well-known models for entity classification in KGs. For the four smaller datasets, these baselines comprise both GCN-based models, such as R-GCN, \textit{E}-R-GCN and CompGCN, 
and embedding-based models including RDF2Vec, ConnectE, and ASSET. 
In addition, we include Feat~\cite{feat}, which uses hand-designed feature extractors, and WL~\cite{WL1,WL2}, which uses graph kernels that count substructures in graphs. For all baselines, we used the recommended hyperparameter settings. As reported in previous work~\cite{rgcn,rdf2vec,rdf2vec_application}, a linear SVM was used to classify the entities using RDF2Vec, WL, ConnectE, and ASSET. 
On the larger datasets, we also compare our approach with both GCN-based and embedding-based models. 
For the GCN-based models, we consider HMGCN, RACE2T, E-R-GCN, RGCN and CompGCN, with R-GCN and CompGCN both using Binary Cross Entropy (BCE) loss. 
The baselines for the embedding-based models consist of ETE~\cite{moon2017learning}, ConnectE, RDF2Vec, and ASSET. 
We reuse the hyperparameters for CompGCN, R-GCN, and \textit{E}-R-GCN as suggested by their respective authors.  

\subsection{Accuracy Results}
\label{sec:accuracyresults}

\textbf{Results on Small Datasets. }
These datasets are designed for single-label classification, i.e., every entity belongs to one class.  
The hyperparameters used for \approach across the different datasets can be found in the GitHub repository.
The results for entity classification are shown in Table~\ref{tab:type_prediction}. 
We see that MPERL+R-GCN outperforms all the baseline methods. 
First, we analyse the performance of MPERL+R-GCN with respect to other GNN-based methods. 
Compared to R-GCN, MPERL+R-GCN does not use a softmax function but a probabilistic loss function consisting of two parts.
As a result, our approach's performance is higher than R-GCN, as the loss used in MPERL+R-GCN captures the information loss between ground truth and predicted distribution. Even though \textit{E}-R-GCN uses the concept of an evidential loss function as well, the end-to-end learning of R-GCN within a Markov process demonstrates a lower sensitivity to noisy neighbors due to the aggregation of the hidden features of each step in the Markov process, as well as a faster convergence over epochs due to the reuse of the hidden feature $h_{n-1}$ in step $n$. 

\begin{table}[t!]
\setlength{\tabcolsep}{1pt}
\centering
\scriptsize
  \caption{Effectiveness results on small datasets} 
  \label{tab:type_prediction}
\begin{tabular}{p{1.2cm}ccccp{0.2cm}ccp{0.6cm}c}\toprule
\textbf{Metrics} & \multicolumn{4}{c}{\textbf{Accuracy}} & \phantom{abc}& \multicolumn{4}{c}{\textbf{F1-Macro Score}}\\ \cmidrule{2-5} \cmidrule{7-10}
\textbf{Dataset} & AIFB & MUTAG & BGS  & AM && AIFB & MUTAG &  BGS & AM \\ \hline
\multicolumn{10}{c}{\textit{Embedding-based methods}} \\ \hline
ConnectE & 83.33 & 75.00 & 79.31 & 88.38 && 80.36 & 70.32 & 74.11 & 86.94\\
RDF2Vec & 88.88 & 67.20 & \underline{87.24} & 88.33 && 86.72 & 62.19 & \underline{85.54} & 87.63\\
ASSET & 86.11 & 76.47 & 75.86 & 89.39 && 82.28 & 67.96 & 73.46 & 87.20\\
\hline
\multicolumn{10}{c}{\textit{Graph featurization methods}} \\ \hline
Feat & 55.55 & 77.94 & 72.41 & 66.66 && 51.69 & 75.82 & 70.88 & 64.87 \\
WL & 80.55 & \textbf{80.88} & 86.20 & 87.37 && 79.43 & \underline{78.52} & 84.89 & 85.85\\
\hline
\multicolumn{10}{c}{\textit{GNN-based methods}} \\ \hline
CompGCN & 86.39 & 66.32& 75.17 & 33.08 && 82.89 & 64.38 & 73.65 & 13.35\\
\textit{E}-R-GCN & \underline{95.56} & 74.56 & 76.55 & \underline{89.85} && \underline{93.21} & 69.63 & 73.98 & \underline{88.76}\\
R-GCN & 92.22 & 73.97 & 75.86 & 89.14 && 87.51 & 70.82 & 74.11 & 79.01\\
\hline
\multicolumn{10}{c}{\textit{Our approach}} \\ \hline
\parbox{1cm}{ \quad \approach \\ +R-GCN} & \textbf{97.22} & \textbf{80.88} & \textbf{89.66} & \textbf{90.40} && \textbf{96.13} & \textbf{79.26} & \textbf{88.26} & \textbf{89.07}\\
\bottomrule
\end{tabular}
\end{table}

Compared to other methods, WL performs well, especially on MUTAG, which matches the highest accuracy achieved by \approach. 
WL also performs well on BGS and AM, making it one of the stronger non-GNN methods. 
From the embeddings-based method, RDF2Vec performs better in terms of the F1-Macro score than other embeddings. 
When comparing results across datasets, the embeddings- and GNN-based perform worse for the MUTAG dataset. 
MUTAG is relatively smaller than other datasets (e.g., BGS and AM) in terms of number of entities, relations, and classes (cf. Table~\ref{tab:dataset}). 
The approaches' performance indicates that learning effective representations for entities in MUTAG is difficult since the connectivity of the entities is rather irregular, as shown in the degree distribution in Figure~\ref{fig:mutag}.
These results show that even in smaller datasets, entity classification can be challenging for state-of-the-art methods.

\begin{table}[t!]
\caption{Effectiveness results on large datasets}
\begin{center}
\scriptsize
\setlength{\tabcolsep}{1pt}
    \begin{tabular}{lrrrrp{0.2cm}rrrr}
        \toprule
        \textbf{Datasets} & \multicolumn{4}{c}{{FB15kET}} &  &\multicolumn{4}{c}{{YAGO43kET}} \\ 
        \cmidrule{2-5} \cmidrule{7-10}
        \textbf{Metrics} & \textbf{MRR} & \textbf{Hit@1} & \textbf{Hit@3} & \textbf{Hit@10} & & \textbf{MRR} & \textbf{Hit@1} & \textbf{Hit@3} & \textbf{Hit@10} \\ \hline
        
        \multicolumn{10}{c}{\textit{Embedding-based methods}} \\ \hline
        ETE & 50.00 & 38.51 & 55.33 & 71.93 & &  23.00 & 13.73 & 26.28 & 42.18 \\
        ConnectE & 59.00 & 49.55 & 64.32 & 79.92 & & 28.00 & 16.01 & 30.85 & 47.92 \\
        RDF2Vec & 59.94 & 50.83 & 64.75 & 77.68 & & \underline{32.74} & \underline{24.17} & \underline{35.94} & \underline{49.76} \\
        ASSET & 60.73 & 51.65 & 65.43 & 78.72 & & 28.11 & 21.13 & 29.93 & 41.53 \\ \hline
        \multicolumn{10}{c}{\textit{GNN-based methods}} \\ \hline
        HMGCN & 51.03 & 39.12 & 54.83 &  72.42 & & 25.01 & 14.19 & 27.33 & 43.68 \\
        RACE2T & 64.14 & 55.56 & 68.40 & \underline{81.36} & & \textbf{34.12} & \textbf{25.27} & \textbf{37.36} & \textbf{52.29} \\
        \textit{E}-R-GCN & \underline{64.59} & \underline{56.94} & \underline{69.12} & 80.09 & & 29.67 &  22.63 & 32.43 & 43.51 \\ 
        R-GCN  & 63.50 & 53.74 & 69.00 & \textbf{82.23} & & 31.56 & 23.32 & 34.53 & 47.49 \\ \hline
        \multicolumn{10}{c}{\textit{Our approach}} \\ \hline
        \parbox{1cm}{ \quad \approach \\ +R-GCN }   & \textbf{65.74} & \textbf{58.18} & \textbf{70.32} &  80.39 & & 30.72  & 23.67 & 33.25 & 43.98  \\
        \bottomrule
    \end{tabular}
    \end{center}
    \label{tab:rgcp_results_all}
\end{table}

\smallbreak
\noindent
\textbf{Results on Large Datasets. }
Next, we assess the performance of our studied approach on 
commonly used large KGs to mimic real-world scenarios, such as YAGO~\cite{suchanek2007yago}, Freebase~\cite{bollacker2008freebase}, in which each entity can have multiple classes but some of which may be missing. 
Therefore, in this study, entity classification corresponds to a multi-label classification problem.  
This task is more challenging than single-label classification, as it requires handling multiple labels for each entity rather than assigning just a single type.
Both benchmarks include a significantly higher number of types and labeled entities than those in single-label entity classification, leading to potential GPU memory problems during our experiments. 
To mitigate this problem, we restrict the maximum number of Markov steps to 2 and apply partial neighborhood sampling. % during training. 
This sampling strategy randomly selects a subset of neighbors for a given entity during message passing in graph neural networks, speeding up training and preventing overfitting. However, it may risk performance degradation if important featured neighbors are not sampled. 
In practice, we only conduct neighbor sampling during training, while all neighbors of the entity are used during inference.
The graphs of large datasets are also augmented with type triples $(e, \textit{hastype}, t_e)$ when training embeddings, as proposed by \cite{pan2021context}, which increases the prediction accuracy.

Table~\ref{tab:rgcp_results_all} shows the performance of our model and the results of the baselines for both benchmarks. 
For the FB15kET dataset, MPERL+R-GCN achieves competitive results and especially outperforms all baselines in terms of the Hit@1 metric, indicating its higher accuracy in the top prediction of the missing types. We observe that MPERL+R-GCN shows significant gains in prediction performance compared to its fundamental model, R-GCN, and slightly performs better than \textit{E}-R-GCN, showing the usefulness of Markov steps in our proposed model. 
However, regarding YAGO43kET datasets, the performance of MPERL+R-GCN is less powerful compared to the FB15kET dataset. 
This discrepancy may arise due to key differences between the benchmarks.
First, as shown in Figure~\ref{fig:degrees}, YAGO contains more higher-degree hub entities (with degrees exceeding $10^4$), which distorts information aggregated from neighbors and reduces model performance, thereby decreasing the performance of our model.
Second, as highlighted in Table~\ref{tab:dataset}, YAGO43kET contains approximately 12 times more classes than FB15kET, further amplifying the decline in performance.
Additionally, due to limited GPU memory, the batch size is set to a small number (16 in practice) for the YAGO43kET dataset, which may potentially lead \approach to converge to sub-optimal solutions.
Overall, MPERL+R-GCN consistently outperforms \textit{E}-R-GCN in both benchmarks. This demonstrates the benefits of incorporating the Markov process, which can reduce sensitivity to noisy neighbors due to the aggregation of hidden features at each Markov step. The results also show that GNN-based methods tend to yield superior outcomes overall compared to embedding-based approaches. These observations show the potential of graph neural networks as a promising technique for addressing entity-type prediction challenges.
\begin{figure}[t!]
\centering
\includegraphics[width=0.5\textwidth]{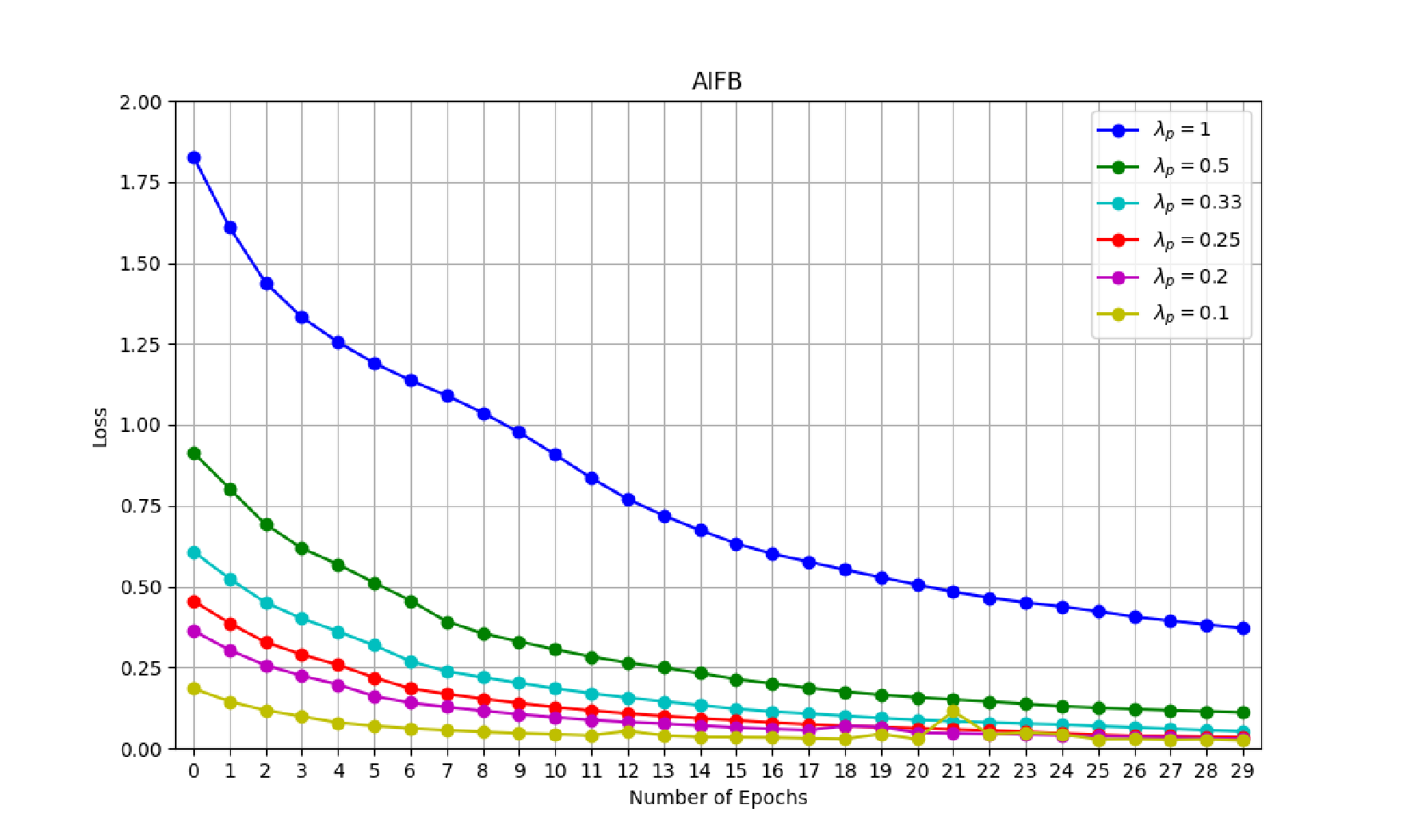}
\caption{Learning curves for the AIFB dataset for different $\lambda_p$ values}
\label{fig:lambda_p_aifb}
\end{figure}

In summary, \approach outperforms the state-of-the-art in small datasets. 
The consistently high F1-Macro scores indicate that \approach can effectively classify entities that belong to least represented classes.
In large datasets, \approach showed very good performance in FB15kET but not in YAGO43kET. 
The sampling techniques implemented to scale \approach to large datasets may have affected the learning process. 
These results suggest that our proposed solution is more suitable for smaller knowledge graphs, where learning meaningful representations is challenging due to the limited information contained in these datasets \textbf{(Q1)}.

\subsection{Impact of the Hyperparameter $\lambda_p$}
\label{sec:iterations}

\begin{figure}[t!]
\centering
\includegraphics[width=0.5\textwidth]{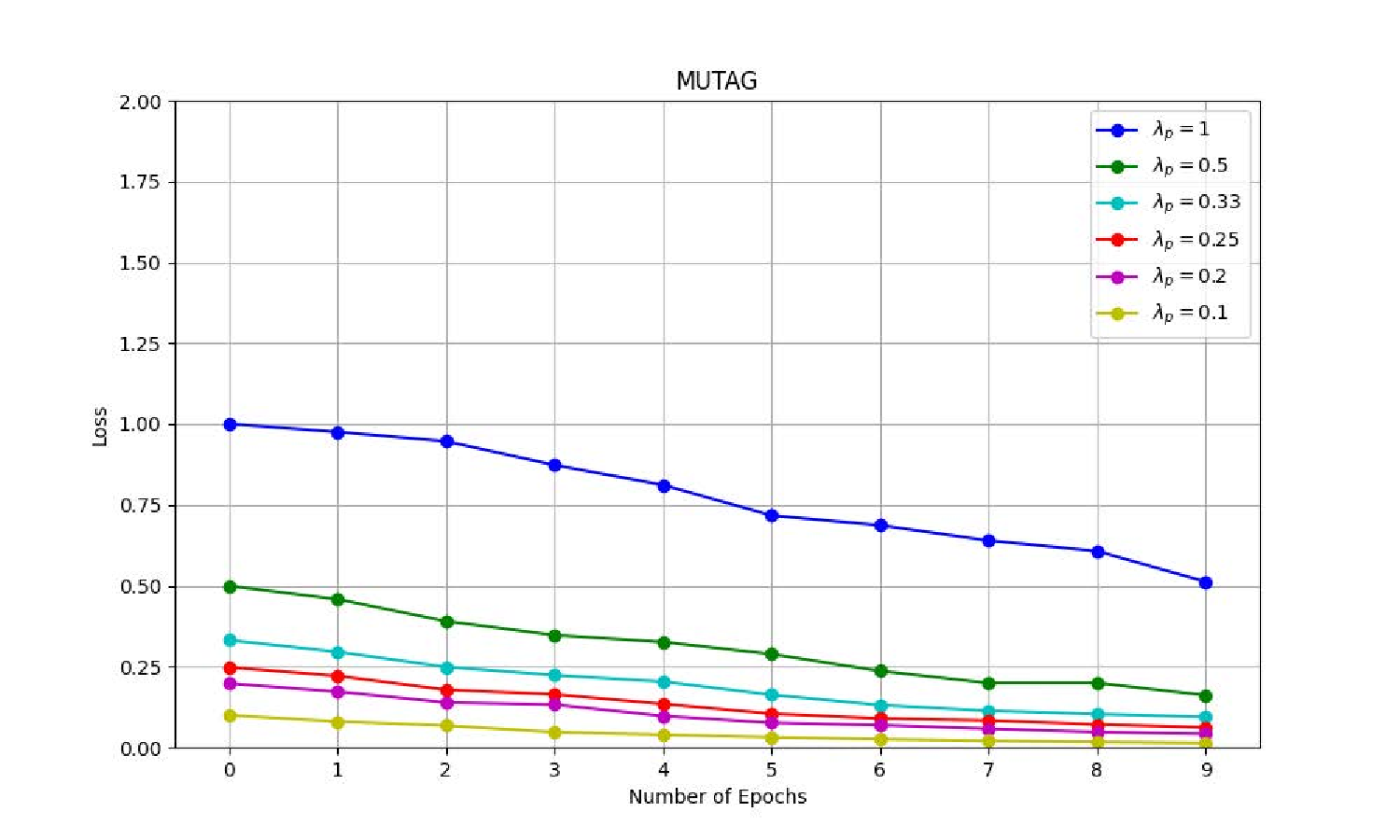}
\caption{Learning curves for the MUTAG dataset for different $\lambda_p$ values}
\label{fig:lambda_p_mutag}
\end{figure}

In the following, we study the impact of the hyperparameter $\lambda_p \in (0,1]$ on the learning process of our approach \approach on selected datasets. 
For this study, we chose the two smallest datasets --  AIFB and MUTAG -- to ensure feasibility, as the study requires training the model multiple times with different $\lambda_p$ values. 
Furthermore, MUTAG is an interesting dataset as the performance of WL was very close to that of \approach, providing meaningful insights.
$\lambda_p$ defines the geometric prior probability distribution $p_{G}(\lambda_{p})$ (see Eq.~\ref{eq:geometric}), which describes the probability that the model enters the absorbing state ($\Lambda_n=1$) in step $n$. The expected value of the distribution $p_{G}(\lambda_{p})$ is $\frac{1}{\lambda_p}$. 
On the one hand, setting it to a low value ensures a high number of steps in the Markov process. 
This strategy encourages higher pondering of the model, yet also leads to increased training time due to the higher number of Markov steps and could result in higher variance.
On the other hand, if $\lambda_p$ is high, it would lead the model to perform only a few Markov steps. 
This setting leads to a reduced training time and should be used to reduce a possible variance. In the special case of $\lambda_p=1$, one Markov step is performed, and therefore, the model corresponds to the evidence-based approach of \textit{E}-R-GCN. 
To study this in more detail, we considered this special case and compared the learning curves between $\lambda_p =1.0$ and smaller values of $\lambda_p$, which correspond to several Markov steps. Figures~\ref{fig:lambda_p_aifb} and \ref{fig:lambda_p_mutag} show the average learning curves of ten runs on selected datasets.

On both datasets, the loss for a single Markov step ($\lambda_p = 1.0$) is initially higher than for smaller values of $\lambda_p$, including $\lambda_p = 0.5$ or two Markov steps. 
This is because the learned features of the previous Markov step are reused in the next one, and the prediction is based on a conjugate prior distribution, which consists of an aggregation of multiple hidden features from each Markov step. This results in much more accurate predictions and, thus, a lower loss.
The learning curve for all $\lambda_p$ values decreases on all datasets over the epochs.
Yet, the loss of the model using $\lambda_p=1.0$ does not drop below that of the model using smaller $\lambda_p$ values. 
Lastly, the difference between $\lambda_p=0.2$ and $\lambda_p=0.1$ becomes almost negligible as the number of Epochs increases in both datasets.

Thus, we can conclude that (i) a higher number of Markov steps generally has a positive effect on the performance of the model compared to a single Markov step, and (ii) a significant decrease of the hyperparameter $\lambda_p$ to $0.1$ -- which produces a significant increase of the number of Markov steps -- does not have a major impact on the performance of the benchmark datasets used as the number of epochs increases \textbf{(Q2)}. 
We therefore recommend $\lambda_p=0.2$ as the default value, representing a trade-off between accurate entity classification and reduced training time achieved with a low number of Markov steps.

\subsection{Ablation Study}
\label{sec:ablation}
\input{tables/ablation_study}

We conduct an ablation study to measure the impact of each component in \approach on the performance.    
By systematically isolating, and evaluating each component of our approach, our ablation study aims to highlight the specific contributions and overall effectiveness of our proposed approach.
The two building blocks of our approach are (i) Markov Learning Process (MP)~(\S\ref{sec:ponder}) and (ii) Evidential Regularized Loss function (ERL)~(\S\ref{sec:loss}). 
When a component is not included, it is represented in the results by a `-', 
e.g., `-ERL' equals the usage of the cross-entropy loss function and `-MP' equals no Markov Process used. 
Conversely, when the component is included, it is represented as `+'. 
The settings for running the experiments are the ones used in \S\ref{sec:accuracyresults}.
For the same reasons explained in \S\ref{sec:iterations}, we selected the datasets AIFB and MUTAG.
The evaluation metrics as reported in Table~\ref{tab:ablationStudyResult}, i.e., accuracy and F1-score, were used.
Including MP alone (+MP, -ERL) decreases the performance significantly for the AIFB dataset, 
yet for MUTAG dataset it leads to a slight increase in accuracy ($+ 1.47$, from $66.18$ to $67.65$)  and in F1-Macro score ($+4.06$, from $43.63$ to $47.69$).
These results indicate that MP alone is not so effective without ERL, and can even be detrimental in some datasets. 
Including ERL alone (-MP, +ERL) does not change the results for accuracy and F1-score in  AIFB. 
In MUTAG, +ERL slightly improves the F1-score ($+6.36$, from $43.63$ to $49.99$) but decreases accuracy ($-2.9$, from $66.18$ to $63.23$); this suggests that +ERL may contribute to better classification of the entities for smaller classes (containing lower number of entities) in MUTAG. 
Still, ERL's impact is minimal without MP, and its effect depends on the dataset. 
Lastly, in both datasets, the results show that including both components (+MP, +ERL) improves the performance significantly and consistently to the highest values, showing that the combined effect of MP and ERL is beneficial \textbf{(Q3)}.

%% file: tables/ablation_study.tex
%\begin{table}[t!]
%\caption{Ablation Study on AIFB and MUTAG datasets}
%\begin{center}
%\footnotesize
%\setlength{\tabcolsep}{1pt}
%    \begin{tabular}{lccp{0.3cm}ccc}
%    \toprule
%    \textbf{Metrics} & \multicolumn{2}{c}{Accuracy} && \multicolumn{2}{c}{F1-Macro Score}\\
%    \cmidrule(lr){2-3} \cmidrule(lr){5-6}
%    \textbf{Dataset} & AIFB & MUTAG && AIFB & MUTAG \\ 
%    \midrule
%    (-MP,-ERL) &  94.44&    66.18       &&   95.08    &   43.63   \\
%    (-MP,+ERL) &  94.44&  63.23        &&   95.08   &    49.99    \\
%    (+MP,-ERL) &  83.33  &    67.65     &&   80.12    &   47.69         \\
%    (+MP,+ERL) &  \textbf{97.22} &    \textbf{80.88}          &&  \textbf{96.13}  &   \textbf{79.26}          \\
%    \bottomrule
%  \end{tabular}
%    \end{center}
%    \label{tab:ablationStudyResult}
%\end{table}

\begin{table}[t!]
\caption{Ablation study on AIFB and MUTAG datasets}
\begin{center}
\footnotesize
\setlength{\tabcolsep}{1pt}
    \begin{tabular}{lccp{0.2cm}ccc}
    \toprule
    \textbf{Dataset} & \multicolumn{2}{c}{AIFB} && \multicolumn{2}{c}{MUTAG}\\
    \cmidrule(lr){2-3} \cmidrule(lr){5-6}
    \textbf{Metrics} & Accuracy & F1-Score && Accuracy & F1-Score \\ 
    \midrule
    (-MP,-ERL) &  94.44&    95.08       &&   66.18    &   43.63   \\
    (-MP,+ERL) &  94.44&  95.08        &&   63.23   &    49.99    \\
    (+MP,-ERL) &  83.33  &    80.12     &&   67.65    &   47.69         \\
    (+MP,+ERL) &  \textbf{97.22} &  \textbf{96.13}  &&  \textbf{80.88}  &   \textbf{79.26}          \\
    \bottomrule
  \end{tabular}
    \end{center}
    \label{tab:ablationStudyResult}
\end{table}

%% file: tex/conclusion.tex
%\vspace{-10mm}
\section{Conclusion and Future Work}
\label{sec:conclusions}

In this work, we have shown how to combine Graph Convolutional Neural Networks (GCN), concepts from Evidential Learning,  and PonderNet in the \approach approach for entity classification in knowledge graphs. 
The model is learned within a Markov decision process, which computes halting at the current step. 
We implemented \approach on top of R-GCN, a state-of-the-art GCN-based architecture. 
The experimental results show a performance increase in most datasets, particularly compared to the previous GCN-based models. 
The aggregation of the hidden features of the individual steps in the Markov process demonstrated a lower sensitivity to noisy neighbors and improved performance. 
The hyperparameter $\lambda_p$ defines the geometric prior probability distribution and, thus, approximates the number of Markov steps. A low $\lambda_p$ setting leads to a high number of Markov steps and, therefore, to a high training time, which has a positive effect on the performance compared to a model that was learned within a Markov process with only one step. The experiments have shown that $\lambda_p=0.2$ (i.e., five Markov steps) has already led to improved performance.
The reuse of the learned hidden features of the previous hidden features of the Markov process has allowed faster convergence of parameters during learning.

As future work, we want to further investigate the hyperparameter $\lambda_p$, by employing an automated approach to select the optimal values.
To improve the scalability of our model in multi-label settings, we also want to replace the Dirichlet distribution with a more suitable alternative, such as the Beta distribution~\cite{zhao2023open}.
Lastly, we want to extend \approach with a gated unit that automatically adjusts the weight of the hidden representation from the previous Markov step ($h_{n-1}$) as input during learning.